\documentclass[journal]{IEEEtran}

\usepackage{amsmath,amssymb}
\usepackage{graphicx}
\usepackage{xcolor}
\usepackage{booktabs}
\usepackage{url}
\usepackage{cite}
\usepackage{tikz}
\usepackage{siunitx}
\usetikzlibrary{arrows.meta,calc,fit,backgrounds}

\newcommand{\Kzero}{K_{0}}
\newcommand{\Srat}{\rho}
\newcommand{\Jac}{J}
\newcommand{\qerr}{\tilde{q}}
\newcommand{\fictheta}{\theta}
\newcommand{\torque}{\tau}
\newcommand{\wrench}{\mathbf{w}}
\newcommand{\gvec}{g}

\begin{document}

\title{A Unified Fractal Impedance Control Framework}

\author{Adnan~Takriti, Mahmoud~Sadeq, Quentin~Rouxel, Daniele~Cafolla,
Michael~Mistry, and~Carlo~Tiseo%
\thanks{A.~Takriti, M.~Sadeq, and C.~Tiseo are with the School of Engineering
and Informatics, University of Sussex, Brighton, UK (e-mail:a.takriti@sussex.ac.uk, c.tiseo@sussex.ac.uk). Quentin Rouxel is with the Department of Mechanical and Automation Engineering, Chinese University of Hong Kong, HK (e-mail:
quentin.rouxel@cuhk.edu.hk). Daniele Cafolla is with the Department of Computer Science, Swansea University, Swansea, UK (e-mail: daniele.cafolla@swansea.ac.uk). Michael Mistry is with School of Informatics, University of Edinburgh, Edinburgh, UK (email:mmistry@ed.ac.uk).}%
\thanks{This work was partially funded by the Royal Society International Exchanges 2025 Cost Share (Italy) scheme under grant number IEC/R2/252022 (``Development of a Biomimetic Synovial Ball and Socket for Robotics'').}
}

\markboth{}%
{Takriti \MakeLowercase{\textit{et al.}}: A Unified Fractal Impedance Control Framework}
\maketitle
\begin{abstract}
Physical human--robot interaction requires controllers capable of maintaining intrinsically bounded interactions with the environment. The Fractal Impedance Controller (FIC) has been proven to guarantee these safety characteristics by design. This paper introduces the robotic implementation and experimental validation of its most advanced architecture: the Harmonic-Passive Motion Paradigm (H-PMP), an evolution of the Passive Motion Paradigm designed to mimic human-like interaction. Furthermore, we present a unified framework for FIC deployment, supported by an open-source software suite featuring ROS2, MuJoCo, and physical hardware integrations. Through three experiments conducted on a Kinova Gen3 arm (trajectory tracking, a wall press with sudden loss of contact, and learning-by-demonstration) we compare the H-PMP against earlier FIC architectures. The results demonstrate that the H-PMP achieves a seamless trade-off between admittance and impedance behaviours. This makes it ideal for deployment in unstructured environments, enabling the robot to transition between interaction modes with no computational overhead.
\end{abstract}

\section{Introduction}\label{sec:intro}

Deploying robots in real-world settings requires the ability to interact
safely with unstructured surroundings. Under these conditions, the
environmental state is notoriously difficult to track and often impossible
to model accurately. Over the past four decades, interaction control has
evolved to manage both expected and unexpected physical contacts, ensuring
safe human-robot and robot-environment
interactions~\cite{hogan1985,haddadin2017,babarahmati2022,wen2023bimanual,tiseo2024,haddadin2024}.
Until recently, these systems were largely confined to research
laboratories and highly specialised industrial contexts, where structured
environments and expert operators ensured proper deployment. However, there
is currently a strong push to introduce robots into unstructured domains
where operators may lack deep technical expertise. Consequently, learning
by demonstration (LbD) and, more recently, generative and agentic AI have
gained traction to bridge the skill gap in robot programming. Despite their
potential, the lack of robustness and explainability in these methods
raises significant safety concerns regarding their deployment near humans.
Furthermore, the extensive data and computational requirements of these
approaches raise critical questions regarding their environmental
sustainability and ethics.

\begin{figure}[!t]
  \centering
  \includegraphics[height=0.285\columnwidth]{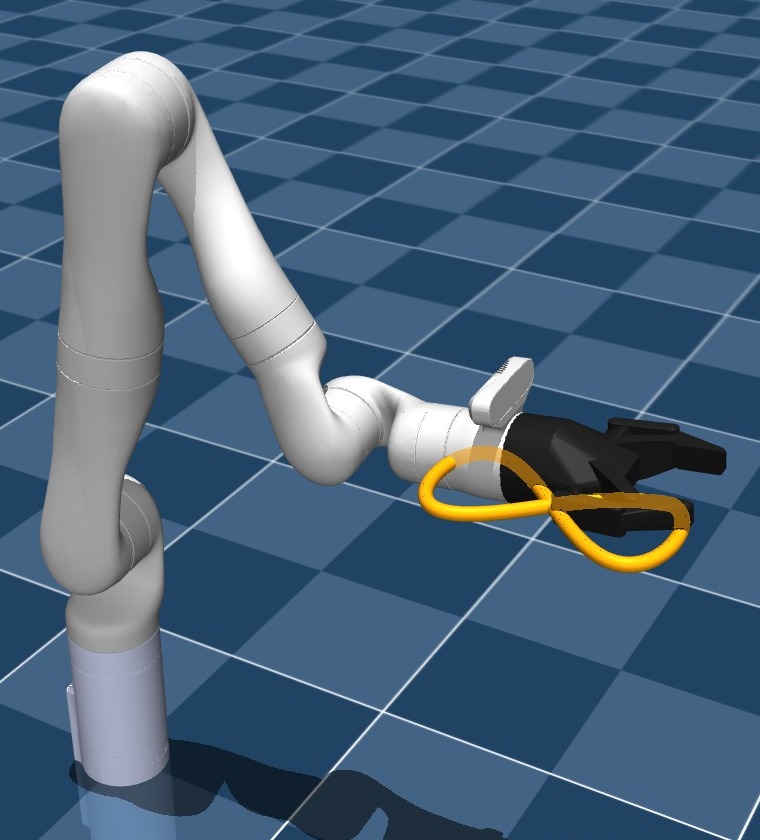}\hspace{0.0025\textwidth}%
  \includegraphics[height=0.285\columnwidth]{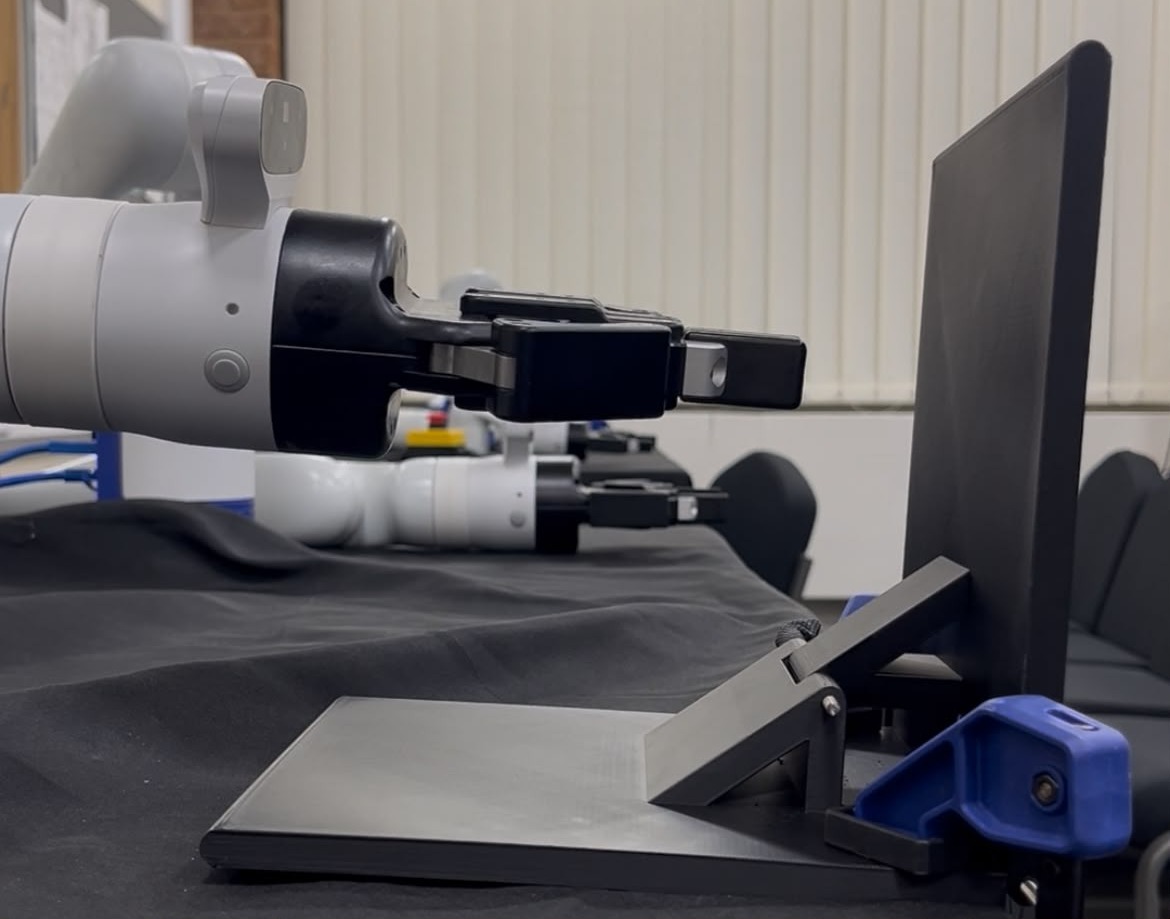}\hspace{0.0025\textwidth}%
  \includegraphics[height=0.285\columnwidth]{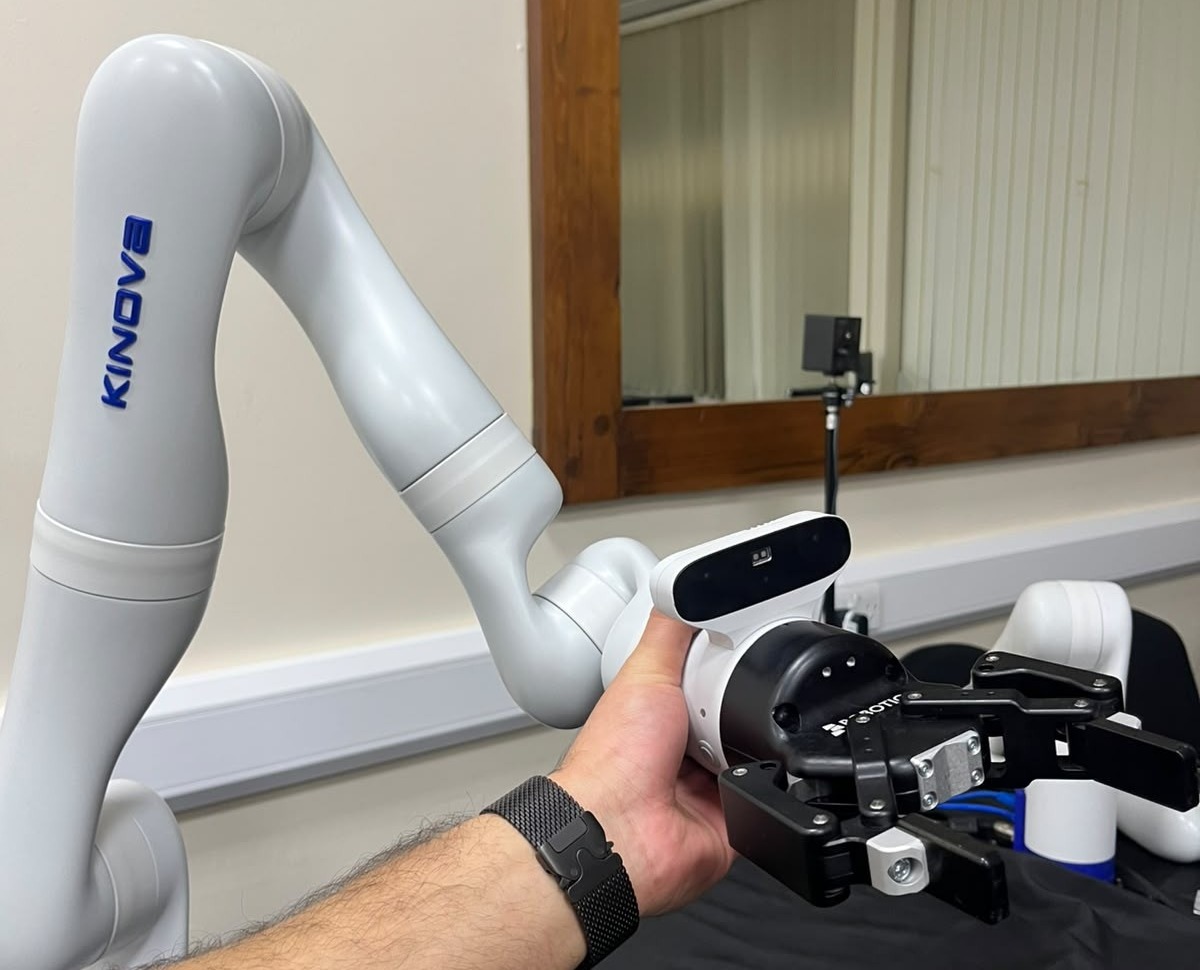}
  \caption{The experiments on the physical arm (Section~\ref{sec:eval}). Left: the lemniscate of the trajectory
    tracking experiment (orange), drawn about the arm in its MuJoCo model, posed from a recorded H-PMP trial; the
    part behind the gripper is shown lighter. Middle: the 7-DOF Kinova Gen3 with its
    gripper closed, facing the wall plate on its quick-release mount (wall press and release). No force sensor
    is fitted: the controllers estimate the external wrench from the joint torques. Right: the operator teaches
    the arm a path by guiding it at the wrist (learning by demonstration).}
  \label{fig:summary}
\end{figure}

Enhancing interaction control beyond the state-of-the-art to provide hard
safety guarantees and increased closed-loop dexterity is crucial prior to
integrating learning methods. This foundational improvement mitigates
safety risks, reduces reliance on data-driven approaches, and increases
sustainability by lowering computational overhead. At the foundation of
interaction control lie port-Hamiltonian controllers (i.e., impedance and
admittance control). Both approach the robot as an equivalent
mass-spring-damper system, acting as reciprocal formulations to each
other~\cite{hogan1985,ott2010}. Admittance control generates a flow command
from an effort error, while impedance control generates an effort command
from a flow error. These approaches carry opposite limitations: impedance
control excels at regulating position, whereas admittance control regulates
interaction force. Biological interaction control in humans and animals
naturally inspires robotic counterparts due to their ability to
effortlessly trade off between these skills during physical
interaction~\cite{hogan1984,burdet2001,yang2011}. While variable
impedance control has been proposed to mimic this biological adaptability,
the stability challenges associated with these controllers have severely
limited their application in complex interactions and high-dimensional
systems~\cite{hannaford2002,ferraguti2013}.

Fractal Impedance Control (FIC)~\cite{babarahmati2022} was introduced to
overcome these fundamental limitations. The controller relies on an
algorithmic formulation of a Li\'enard equation, which describes non-linear
systems such as Van der Pol oscillators \cite{tiseo2021exploiting}. The system generated by the FIC
acts as a critically damped, asymptotically stable attractor around the
desired state, defined algorithmically from an effort profile \cite{babarahmati2022,tiseo2020biomimetic}.
Consequently, it enables the generation of asymptotically stable, passive
controllers that can be safely superposed and tuned online. FIC has
successfully demonstrated the capability to introduce an outer admittance
loop on top of a non-linear impedance controller, deployed within
biomimetic architectures for robust interaction in challenging dynamic
environments \cite{wen2023bimanual,tiseo2022ik}. Its intrinsically passive behaviour also yields excellent
performance despite significant feedback delays and low control
bandwidths, making it exceptionally well-suited for teleoperation
applications\cite{babarahmati2022,tiseo2024medical,wen2023bimanual,Babarahmati2021}.

This manuscript presents the integration of these theoretical advancements
into a unified library to support the development of FIC-based controllers
for manipulation and teleoperation. Notably, the library includes a robotic
implementation of the Harmonic-Passive Motion Paradigm (H-PMP), the most
advanced FIC architecture developed to model human motor control \cite{tiseo2021hpmp}. H-PMP can
autonomously trade off admittance and impedance behaviours without requiring
a rigid state machine or model-based interaction optimisation \cite{tiseo2021hpmp}. Furthermore,
the library is fully integrated with standard robotics infrastructure,
including MuJoCo for simulation, the Pinocchio library for inverse
kinematics optimisation, and ROS2. To streamline deployment, we introduce
FIC Studio---a user-friendly, modular Graphical User Interface (GUI) that
centralises these functionalities, featuring drag-and-drop controller
tuning and Python live scripting for high-level control integration. Fig.~\ref{fig:summary} shows the arm in two of the experiments for the evaluation of the interaction  performance.

The remainder of this paper is structured as follows:
Section~\ref{sec:related} introduces related work.
Section~\ref{sec:system} provides an overview of the FIC Suite functionalities.
Section~\ref{sec:modes} describes the FIC library and the robotic
implementation of H-PMP. Section~\ref{sec:eval} presents an evaluation
comparing the proposed control architecture against an earlier
implementation of the Task-Space FIC. Finally,
Section~\ref{sec:conclusion} draws the conclusions.

\section{Related Work}\label{sec:related}
As mentioned in the introduction, interaction control aims to develop an
architecture capable of modulating its behaviour to naturally trade off
between admittance and impedance strategies. This type of adaptability has
been sought since the early nineties but was long relegated to
non-redundant
robots~\cite{tiseo2024,mcgrath2023repetitive,tiseo2020biomimetic,hapfic2021}.
In non-redundant systems, the transformation between joint and task spaces
is fully observable; for instance, most actuators equipped with torque
sensors are governed by an impedance controller featuring an outer
admittance loop. However, this does not hold for redundant robots, where
the transformation can only be approximated using null-space projectors,
leading to non-unique solutions. Consequently, interaction control for
redundant robots traditionally had to operate strictly in either admittance
or impedance modes \cite{tiseo2024,wen2023bimanual,ajoudani2012,haddadin2024}. Optimisation-based approaches, such as Model Predictive
Control (MPC), have enabled substantial improvements in implementing
admittance-impedance trade-off strategies. Nevertheless, they introduce
vulnerabilities related to modelling errors (i.e., the sim-to-real gap) and
carry the risk of unstable behaviour stemming from the numerical
instability of the optimisation problem itself \cite{ajoudani2012,haddadin2024,wen2023bimanual,tiseo2024}.

\begin{figure*}[!t]
  \centering
  \begin{tikzpicture}[
    font=\footnotesize,
    >=Stealth,
    box/.style={draw, rounded corners=2pt, align=center, inner sep=2pt,
                minimum height=0.8cm},
    gui/.style={box, fill=orange!12, minimum width=2.65cm, minimum height=2.5cm},
    node/.style={box, fill=blue!9, minimum width=2.5cm},
    ext/.style={box, fill=gray!10, minimum width=3.3cm},
    cfg/.style={box, fill=green!10},
    arr/.style={->, thick},
    biarr/.style={<->, thick},
    lab/.style={font=\footnotesize, align=center, inner sep=1.5pt},
    group/.style={draw=gray, dashed, rounded corners=4pt, inner sep=3pt},
]

\node[gui] (studio) at (0, 0) {\textbf{FIC Studio}\\(desktop / web)\\[2pt]
    dashboard\\live tuning\\scripts, sequences\\robot configuration};

\node[draw, fill=gray!5, rounded corners=2pt, minimum width=1.9cm,
      minimum height=2.5cm, align=center, inner sep=2pt] (ros) at (2.85, 0) {\textbf{ROS~2}\\[2pt]
    topics\\\texttt{robot\{N\}\_*}};

\draw[biarr] (studio.east) -- (studio.east -| ros.west);

\node[node, minimum height=2.5cm] (ctl) at (6.2, 0) {\texttt{fic\_control}\\[2pt]
    task-space layer\\FIC library, 1\,kHz\\[2pt]simulation or\\hardware target};
\node[node, minimum height=2.5cm, minimum width=2.3cm] (il) at (9.55, 0) {\texttt{fic\_inner\_loop}\\[2pt]
    joint layer\\[2pt]hold and ramp};
\draw[arr] (ctl.east) -- node[lab, above]{UDP} (il.west);
\draw[biarr] (ros.east) -- node[lab, above]{commands\\[-2pt]states} (ctl.west);

\node[ext] (mj) at (14.0, 0.95) {\textbf{MuJoCo node}\\[1pt]
    physics, contacts};
\node[ext] (drv) at (14.0, -0.95) {\textbf{Vendor driver and robot}\\[1pt]
    Kortex $\cdot$ libfranka\\UR $\cdot$ FRI $\cdot$ CAN};

\draw[biarr] (il.east |- mj.west) -- node[lab, above]{lockstep} (mj.west);
\draw[biarr] (il.east |- drv.west) -- node[lab, above]{shared}node[lab, below]{memory} (drv.west);

\node[cfg, minimum height=0.55cm] (cfgf) at (6.2, -1.85) {robot (YAML) $\cdot$ profiles (XML) $\cdot$ Pinocchio};
\draw[arr, dashed] (cfgf.north) -- (ctl.south);

\begin{scope}[on background layer]
  \node[group, fit=(mj), label={[lab, font=\footnotesize\itshape, text=blue!55!black, anchor=south east]north east:Simulation}] {};
  \node[group, fit=(drv), label={[lab, font=\footnotesize\itshape, text=black!65, anchor=north east]south east:Hardware}] {};
\end{scope}

\end{tikzpicture}
  \caption{The FIC Suite. FIC Studio acts on the robot only through ROS2
    topics: commands (mode, targets, profile selection, parameter updates)
    and states (joint and end-effector state, wrench, controller
    internals). The robot node, \texttt{fic\_control}, runs the task-space
    layer of the FIC library at \SI{1}{\kilo\hertz} and sends each tick to the
    joint inner loop, \texttt{fic\_inner\_loop}, which steps with MuJoCo in
    simulation or exchanges torques with the vendor driver through shared
    memory on the robot.}
  \label{fig:suite}
\end{figure*}
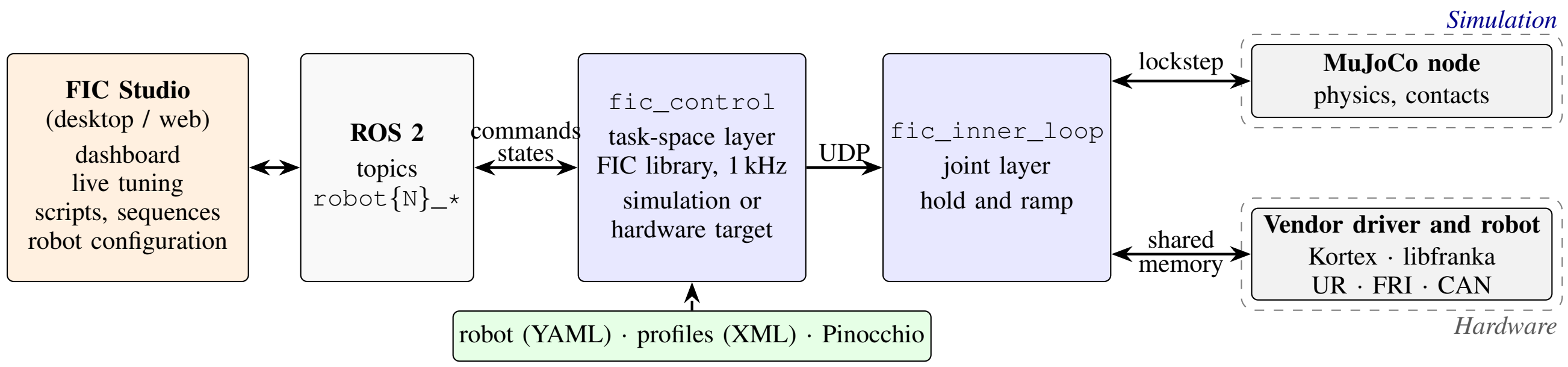

FIC was introduced to solve this problem by exploiting passivity. It is
important to note that FIC did not introduce the concept of the passive
impedance controller, which has existed since its inception in
1985~\cite{hogan1985}. If an impedance controller tracks a zero-velocity
setpoint, it is generally a passive system---and conservative if damping is
fully omitted. However, removing velocity tracking from a linear impedance
controller historically meant the system could no longer accurately track
dynamic trajectories. Prior to FIC, numerous methods attempted to resolve
this limitation by introducing adaptive behaviours, energy budgets, and
hierarchical
optimisations~\cite{ott2010,ferraguti2013,villani2008,dimeas2016,keemink2018,colgate1994,abudakka2020,ajoudani2012,ferraguti2019,michel2022,muehlbauer2026,lachner2021,dietrich2020}.
Yet, tracking the energy of a redundant system is highly non-trivial; even
under ideal conditions with a perfect model, non-passive systems require a
path integral. This inevitably introduces integration drift in the state
observer, leading to subsequent instability. By contrast, passive
controllers hold the distinct advantage of avoiding integration drift
because their state observer is inherently a conservative system.

Here, a critical distinction must be drawn between a passivated system
(e.g., using energy tanks) and a purely passive system~\cite{tiseo2024}. A
passivated system remains an active system that relies on an additional
conservative observer (i.e., a virtual spring) to enforce passivity by
storing the non-conservative energy generated during interaction. While
this approach improves stability and performance, it remains susceptible to
instability or degraded tracking performance~\cite{tiseo2024}.

The FIC was originally envisioned as an online tunable passive controller
that guarantees asymptotic stability by design. Over the years, it has
evolved from a bio-inspired concept designed to mimic human motor control
into a mature architecture capable of addressing major challenges in
interaction control and teleoperation by exploiting its intrinsic
passivity. It operates as a controller capable of leveraging highly
non-linear impedance profiles to maintain excellent performance across
diverse environmental dynamics with minimal tuning, all without requiring
prior knowledge of the system or environmental dynamics. However, its
development has, so far, remained fragmented across multiple
standalone implementations. This work brings these disparate elements
together, packaging them into a single, unified tool to streamline
implementation and integration. Additionally, we port, for the first time
to robotics, the complete Harmonic-Passive Motion Paradigm (H-PMP)
originally developed to model human motor control~\cite{tiseo2021hpmp}.
Previously, this paradigm had only been partially adapted for robotics by
replacing the lower-level control with dynamic
compensation~\cite{wen2023bimanual,tiseo2024medical}.
\section{The FIC Suite}\label{sec:system}

The FIC Suite is cross-platform (compatible with Linux, macOS, and
Windows), and a Docker containerised version is also available. The suite
comprises three main components interfaced via ROS2 nodes: the GUI, the
robot nodes (handling simulation and hardware interfaces), and the core FIC
library. For rigid body kinematics and differential kinematics modelling,
the suite integrates the Pinocchio library~\cite{pinocchio} due to its
state-of-the-art computational efficiency. However, the modular
architecture ensures that users can easily substitute Pinocchio with their
kinematics library of choice.

Fig.~\ref{fig:suite} shows the three components and their interfaces. The
suite runs natively on Linux, with simulation and hardware, and on macOS,
with simulation; the Docker version runs the simulation on Windows, macOS
and Linux, with the GUI in a web browser~\cite{ficinterface}.

\subsection{FIC Studio}\label{sec:system:studio}

The GUI serves as the primary interaction point for the end-user. It
bridges the FIC control libraries with MuJoCo~\cite{mujoco} for simulation
and with physical robots through one driver per vendor interface, e.g., the
Kinova Kortex API, the Franka Control Interface, the Universal Robots
client library, the KUKA Fast Robot Interface and CAN bus motors. A driver
only exchanges joint states and torques with the controller, so a new robot
needs a driver and a configuration file rather than changes to the
controllers. So far, the hardware chain has been tested on the Kinova Gen3
and the Franka FR3. In simulation, the same controllers also drive parallel
mechanisms and dexterous hands. A robot is added by dropping its URDF or
MuJoCo model, from which the Studio writes the missing model and the
robot's configuration (Section~\ref{sec:system:library}). A per-robot
hardware model adds to the simulation the joint friction, damping and
torque-sensor noise estimated on the real arm, so that a tuning carries
over to the hardware. Grippers, hands and tools, including custom ones, are
robot types of their own that can be fitted to an arm after a check against
its payload and joint torque limits. A dedicated
drag-and-drop tuning interface enables users to seamlessly adjust
controller parameters and store these configurations in an XML profile
file. Furthermore, the GUI includes a Python live scripting environment for
implementing custom high-level control functions, as well as an interface
for saving and executing motion sequences. These sequences can be defined
via scripts, inputted manually through the GUI, or recorded directly by
moving the robot (kinesthetic teaching).

\begin{figure}[!t]
  \centering
  {\setlength{\fboxsep}{0pt}\setlength{\fboxrule}{0.4pt}%
  \fbox{\includegraphics[width=0.485\textwidth]{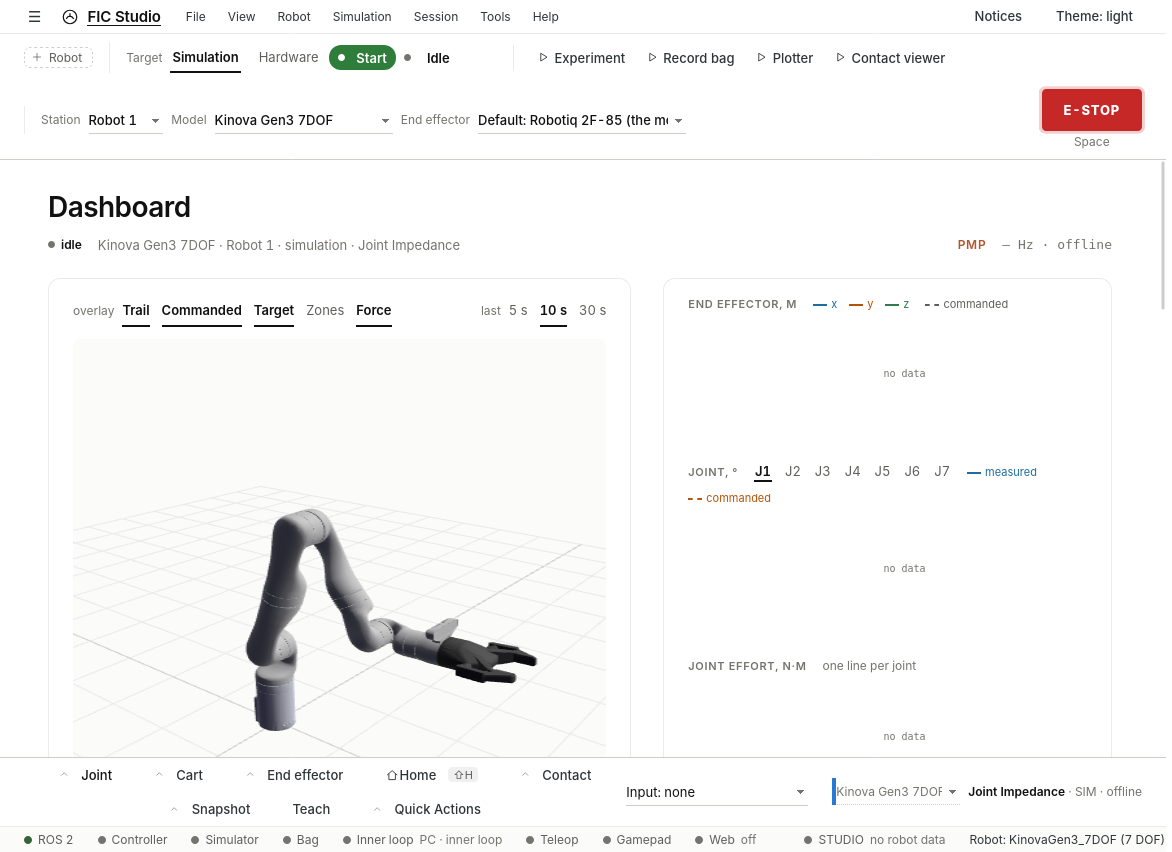}}\hfill
  \fbox{\includegraphics[width=0.485\textwidth]{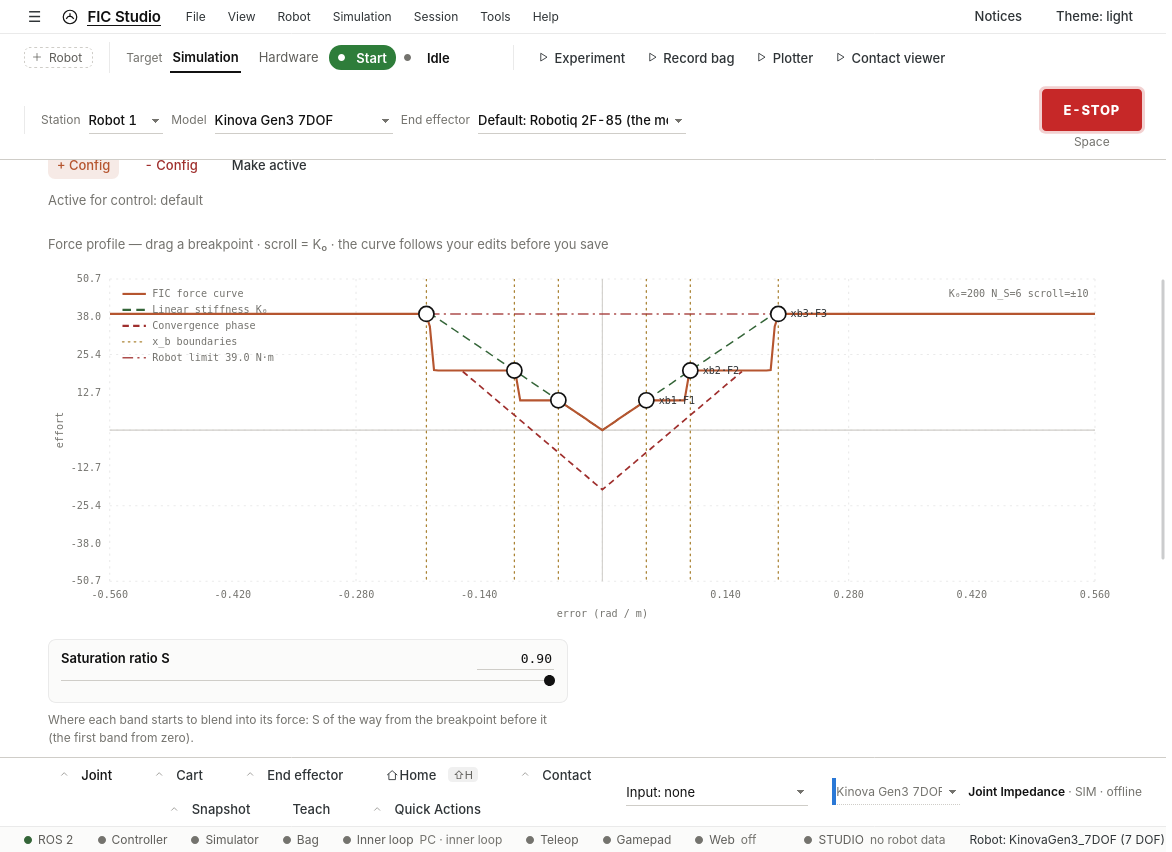}}}
  \caption{FIC Studio's web interface (Kinova Gen3, simulation target). Top:
    the dashboard, with the robot in 3D and live plots of the end-effector
    position, the joint angles and the joint efforts. Bottom: the tuning panel where the user can update the controller effort profiles.}
  \label{fig:studio}
\end{figure}

The GUI, FIC Studio, runs as a desktop window and as a web interface, which
other computers and phones reach after signing in, with an emergency stop
on every page. It selects the robot, the target (simulation or hardware),
the control mode and the active profile. Its dashboard shows the robot in
3D with its trail, commanded path, target, safety zones and contact force,
together with the tracking error and the update rate
(Fig.~\ref{fig:studio}). Pressing Start launches the simulator, or the
robot's vendor driver, and \SI{2}{\second} later the controller; Stop ends
the controller first. Since both targets load the same library and profile
file, a profile tuned in simulation loads on the robot unchanged. The
topics are scoped by a robot number, so one Studio can run several robots, simulated or real, side by side, and the Studios of
a lab find one another on the network, so that one station's arm can be
teleoperated from another.

\paragraph*{Live tuning}\label{sec:system:tuning}
The tuning page draws a channel's force profile (Section~\ref{sec:effort})
as a canvas whose stiffness $\Kzero$, breakpoints $x_{b,j}$, saturation
efforts $F_{s,j}$ and ratio $\Srat$ are set by dragging handles, scrolling and
a slider. The
channel's live operating point is drawn on the curve, and the tracking
error, joint effort and Cartesian wrench are shown beneath it
(Fig.~\ref{fig:studio}). Saving writes the profile file and publishes a parameter update. The robot
node then reads the file off its control thread and passes the new profile
to the inner loop, which switches to it on its next tick, without a restart
or a mode switch. Similarly, the
experiments of Section~\ref{sec:eval} are Studio extensions: Python scripts,
run headless, that act on the robot only through the same ROS2 topics.

\subsection{Robot Nodes and Runtime}
\label{sec:system:interface}\label{sec:system:rt}\label{sec:system:comms}\label{sec:system:kin}

The robot node, \texttt{fic\_control}, runs the task-space layer on a
\SI{1}{\milli\second} wall-clock timer: the FIC control law, the
kinematics, the safety checks and, on a thread of its own, the inverse
kinematics (IK). Each tick, it sends a command over UDP to the joint inner
loop, \texttt{fic\_inner\_loop}, a separate process that runs the joint
modes, H-PMP's joint layer and a hold-and-ramp that stops the arm when the
task layer goes quiet; the same joint layer also runs on an MCU, as in
Experiment~C. Both targets load the same library, profile file and URDF,
that of the arm with its fitted end-effector
(Section~\ref{sec:system:library}); in the experiments, the Kinova Gen3 with
a Robotiq 2F-85 gripper. A target parameter selects the plant. In
simulation, the inner loop steps in lockstep with MuJoCo, and the gravity
torque is the simulator's; on the robot, it runs at \SI{1}{\kilo\hertz}
against the vendor driver through shared memory, the gravity torque comes
from Pinocchio, and the driver adds its own torque limits.

\begin{table}[!t]
\renewcommand{\arraystretch}{1.1}
\caption{FIC Studio Control Modes}
\label{tab:modes}
\centering
\footnotesize
\begin{tabular}{@{}p{0.07\columnwidth}p{0.52\columnwidth}p{0.31\columnwidth}@{}}
\toprule
\textbf{\#} & \textbf{Name} & \textbf{Space} \\
\midrule
1 & Impedance              & Joint \\
2 & Impedance with planner    & Joint \\
3 & Admittance         & Joint \\
4 & HapFIC        & Joint \\
\midrule
5 & Impedance              & Cartesian \\
6 & Impedance with planner    & Cartesian \\
7 & Admittance          & Cartesian \\
8 & HapFIC         & Cartesian \\
\bottomrule
\end{tabular}

\smallskip
\end{table}

\subsection{Library Core and Configuration}\label{sec:system:library}

At its core, the FIC Library is built around a C++ template class,
\texttt{ficControl}. This class contains all the methods required to
implement ad hoc FIC controllers, alongside a collection of standard FIC
architectures that can be called directly. These architectures are the
modes of Section~\ref{sec:modes}. The class owns an indexed array of
per-DOF controllers, planners and admittance integrators, whose channels
are numbered from 0. Joint $i=1,\dots,n \in \mathbb{N}$ uses channel $i-1$, the Cartesian
axes use channels 0--3, and the postural controller and H-PMP's task layer
use channels of their own. Each robot is described by configuration rather
than code. A YAML file per robot gives its URDF and MuJoCo model, the
kinematics solver, the joint limits, the per-joint torque limits $L_i$, the
admittance deadbands, the safety settings and the hardware driver. The
controllers load the URDF of the arm with its end-effector: either the
robot's own, or one that the Studio generates by adding the end-effector's
mass and inertia to the link it is mounted on, so that the gravity
compensation carries it. XML files, one per robot and one per tuning
profile, give the per-mode, per-DOF profiles. The library provides full access to all controller tuning parameters, granting
advanced users complete customisation over the control behaviour. Finally,
the FIC library is highly modular and can be deployed as a standalone
dependency if the broader GUI and ROS2 infrastructure is not required.

\section{FIC Controllers}\label{sec:modes}

Table~\ref{tab:modes} summarises all the controllers available within the
library. The joint-space and task-space admittance controllers, impedance
controllers, and planners are formulated as linearly independent
collections of their respective mono-dimensional functions, which will be
detailed later in this section. The quaternion controller is presented
separately due to the distinctive formulation of the FIC in orientation
space; specifically, the error and torque profiles are defined directly in
quaternion space before being projected onto the end-effector frame.

\subsection{FIC Effort Function}\label{sec:effort}

The FIC formulation requires the effort functions to be smooth and possess
bounded derivatives, thereby ensuring Lipschitz continuity \cite{babarahmati2022,hapfic2021,tiseo2021hpmp}.
The effort profile implemented in the library is linear in the vicinity of
the desired state, featuring $n$ sigmoidal saturations for negative state
errors, $m$ sigmoidal saturations for positive state errors, and $k=n+m$.
Let $\tilde{x}=x_d-x$ denote the state error, where $x_d$ is the desired
state and $x$ is the current state. The effort function is then defined as:
\begin{equation}
\label{eq:fic-static}
\Gamma(\tilde{x})=
\begin{cases}
\nu^{1}(\tilde{x}), & \tilde{x}\le\tilde{x}_{b0}^{1},\\[-2pt]
\ \ \vdots & \\
\nu^{n}(\tilde{x}), & \tilde{x}_{b0}^{n-1}\le\tilde{x}\le\tilde{x}_{b0}^{n},\\[2pt]
K\tilde{x}, & \tilde{x}_{b0}^{n}\le\tilde{x}\le\tilde{x}_{b0}^{n+1},\\[2pt]
\psi^{n+1}(\tilde{x}), & \tilde{x}_{b0}^{n+1}\le\tilde{x}\le\tilde{x}_{b0}^{n+2},\\[-2pt]
\ \ \vdots & \\
\psi^{k}(\tilde{x}), & \tilde{x}\ge\tilde{x}_{b0}^{k},
\end{cases}
\end{equation}
Let $\tilde{x}_{b}^{i} = \tilde{x}_{b0}^{i}/S^i$ denote the point at which the saturation reaches its peak value. Here, $S^i$ represents the percentage of the distance between $\tilde{x}_{b}^{i}$ and the previous transition point in the effort profile (i.e., the origin or a preceding boundary, $\tilde{x}_{b}$) where the effort saturation initiates.
The sigmoidal saturation profiles are then defined as follows:
\begin{equation}
\begin{array}{l}
     \nu^i(\tilde{x})=\left(\gamma^{i}-\gamma^{i+1}\right)\left(\mathit{tanh}\left(\alpha^i+\pi\right)\right)+\gamma^{i+1}\\\\
     \psi^i(\tilde{x})=\left(\gamma^{i+1}-\gamma^{i}\right)\left(\mathit{tanh}\left(\alpha^{i}+\pi\right)\right)+\gamma^{i}\\\\
      \alpha^i=2\pi\frac{x-\left(S^i\left(x_b^i-x_{b0}^i\right)+x_{b0}^i\right)}{\left(1-S^i\right)\left(x_b^i-x_{b0}^i\right)}\\
\end{array}
\end{equation}
where $\gamma_i$ is the $i^{th}$ saturation value of the effort. 
\subsection{FIC Control}\label{sec:atomic}
The FIC is a algorithmic implementation of a critically damped Liénard equation; thus, belonging to the same type of dynamic system as the Van der Pol oscillator \cite{babarahmati2022,tiseo2024,tiseo2021exploiting}. The algorithm generating the FIC attractor is:
\begin{equation}
    \Gamma_c(\tilde{x})=\left\{\begin{array}{ll}
      \Gamma(\tilde{x})   & \text{Divergence}\\\\
        \frac{2\Gamma(\tilde{x}_M)}{\tilde{x}_M}\left(\frac{\tilde{x}_M}{2}-\tilde{x}\right) &   \text{Convergence}
    \end{array} \right.
\end{equation}
where $\tilde{x}_M$ is the maximum state error recorded at the end of the previous divergence phase.

\subsection{FIC Quaternion}\label{sec:quaternion}

Let $q_{\mathrm{des}}$ and $q$ represent the desired and current
end-effector orientations in quaternion form. The orientation error is
defined as $\qerr=q_{\mathrm{des}}\otimes q^{*}$. The angular error
$\fictheta$ in quaternion space can be derived from
$\qerr=[\qerr_s,\qerr_v]^{T}$, where $\qerr_s$ and $\qerr_v$ are the scalar
and vector components of the error quaternion, respectively:
\begin{equation}\label{eq:geo-angle}
\fictheta = 2\,\mathrm{atan}\!\Big(\frac{\lVert\qerr_v\rVert}{\qerr_s}\Big),
\end{equation}
This angular error is then used to compute the torque at the end-effector,
$\torque_{ee}$, as follows:
\begin{equation}\label{eq:ori-torque}
\torque_{ee}=\Gamma_c(\fictheta)\,\frac{\qerr_v}{\lVert\qerr_v\rVert},
\end{equation}
where $\Gamma_c$ is the effort function of Eq.~\eqref{eq:fic-static} with the
parameters of the orientation channel (channel~3), applied through the
control law of Section~\ref{sec:atomic}.

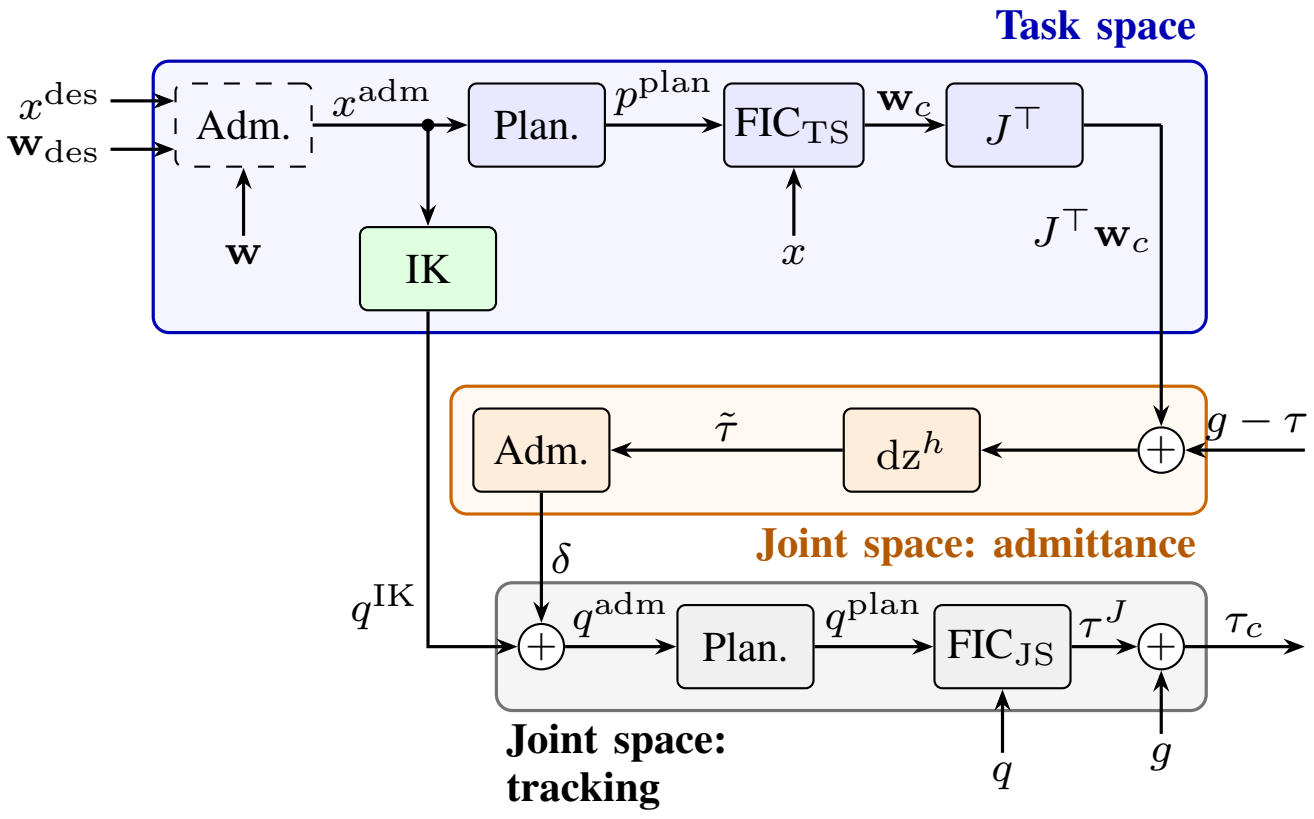
\begin{figure}[!t]
  \centering
  \begin{tikzpicture}[
    font=\footnotesize,
    >={Stealth[length=4pt]},
    box/.style={draw, rounded corners=1.5pt, align=center,
                minimum height=0.55cm, minimum width=0.9cm, inner sep=2pt},
    cart/.style={box, fill=blue!9},
    adm/.style={box, fill=orange!14},
    jnt/.style={box, fill=gray!12},
    ikb/.style={box, fill=green!12},
    opt/.style={box, fill=blue!4, dashed},
    sum/.style={draw, circle, fill=white, inner sep=0pt, minimum size=0.3cm, font=\scriptsize},
    layer/.style={draw=#1, line width=0.6pt, rounded corners=3pt, inner sep=4pt},
    layerlab/.style={font=\footnotesize\bfseries, text=#1!70!black, inner sep=2pt},
    arr/.style={->, semithick},
    line/.style={semithick},
    lab/.style={font=\footnotesize, inner sep=1.5pt},
]

% Task space
\node[opt]  (cadm) at (0.78,0)  {Adm.};
\node[cart] (plan) at (2.72,0)  {Plan.};
\node[cart] (imp)  at (4.42,0)  {FIC$_{\mathrm{TS}}$};
\node[cart] (jt)   at (5.88,0)  {$\Jac^{\top}$};
\coordinate (xa) at (2.0,0);
\node[ikb]  (ik)   at (2.0,-0.95) {IK};
\coordinate (cR) at (6.85,0);

\draw[arr] (-0.1,0.16) node[lab, left]{$x^{\mathrm{des}}$} -- (cadm.west |- 0,0.16);
\draw[arr] (-0.1,-0.16) node[lab, left]{$\wrench_{\mathrm{des}}$} -- (cadm.west |- 0,-0.16);
\draw[arr] ($(cadm.south)+(0,-0.45)$) node[lab, below]{$\wrench$} -- (cadm.south);
\draw[arr] (cadm.east) -- node[lab, above, pos=0.45]{$x^{\mathrm{adm}}$} (plan.west);
\fill (xa) circle (1.2pt);
\draw[arr] (xa) -- (ik.north);
\draw[arr] (plan) -- node[lab, above]{$p^{\mathrm{plan}}$} (imp);
\draw[arr] ($(imp.south)+(0,-0.45)$) node[lab, below]{$x$} -- (imp.south);
\draw[arr] (imp) -- node[lab, above]{$\wrench_{c}$} (jt);

% Joint space: admittance
\node[sum] (s1) at (6.85,-2.15) {$+$};
\node[adm] (db) at (5.2,-2.15)  {$\mathrm{dz}^{h}$};
\node[adm] (m2) at (2.75,-2.15)  {Adm.};
\draw[line] (jt.east) -- (cR);
\draw[arr] (cR) -- node[lab, left, pos=0.35]{$\Jac^{\top}\wrench_{c}$} (s1.north);
\draw[arr] (7.8,-2.15) -- node[lab, above, pos=0.4]{$\gvec-\torque$} (s1.east);
\draw[arr] (s1.west) -- (db.east);
\draw[arr] (db.west) -- node[lab, above]{$\tilde{\torque}$} (m2.east);

% Joint space: tracking
\node[sum] (s2) at (2.75,-3.45)  {$+$};
\node[jnt] (jp) at (4.1,-3.45)  {Plan.};
\node[jnt] (jf) at (5.8,-3.45) {FIC$_{\mathrm{JS}}$};
\node[sum] (s3) at (6.85,-3.45) {$+$};
\draw[arr] (ik.south) |- node[lab, left, pos=0.44]{$q^{\mathrm{IK}}$} (s2.west);
\draw[arr] (m2.south) -- node[lab, right]{$\delta$} (s2.north);
\draw[arr] (s2.east) -- node[lab, above]{$q^{\mathrm{adm}}$} (jp.west);
\draw[arr] (jp) -- node[lab, above]{$q^{\mathrm{plan}}$} (jf);
\coordinate (qin) at ($(jf.south)+(0,-0.42)$);
\draw[arr] (qin) node[lab, below]{$q$} -- (jf.south);
\draw[arr] (jf.east) -- node[lab, above]{$\torque^{J}$} (s3.west);
\coordinate (gin) at ($(s3.south)+(0,-0.42)$);
\draw[arr] (gin) node[lab, below]{$\gvec$} -- (s3.south);
\draw[arr] (s3.east) -- node[lab, above]{$\torque_{c}$} (7.8,-3.45);

\begin{scope}[on background layer]
  \node[layer=blue!70!black, fill=blue!4, fit=(cadm)(plan)(imp)(jt)(ik)(s1.east |- cR),
        label={[layerlab=blue, anchor=south east]north east:Task space}] {};
  \node[layer=orange!80!black, fill=orange!5, fit=(m2)(db)(s1),
        label={[layerlab=orange, anchor=north east]south east:Joint space: admittance}] {};
  \node[layer=black!55, fill=black!4, fit=(s2)(jp)(jf)(s3),
        label={[layerlab=black, anchor=north west, align=left]south west:Joint space:\\tracking}] {};
\end{scope}

\end{tikzpicture}
  \caption{The H-PMP control architecture comprises a task-space admittance controller and a joint-space HapFIC. \emph{Task-Space}: The system can implement either a traditional admittance behaviour or a HapFIC by altering how the admittance modifies $x^{\mathrm{des}}$ to yield $x^{\mathrm{adm}}$. This output is then passed to the Cartesian planner (Plan.) and a Task-Space FIC (FIC$_\mathrm{TS}$), which consists of a 3D Cartesian controller and a quaternion controller. The resulting output wrench is transformed into the desired joint torques by pre-multiplying it by the Jacobian transpose ($J^T$). \emph{Joint-Space}: The torque error $\torque_\mathrm{des} - \torque + g(q)$ is processed by the dead-band ($\mathrm{dz}^{h}$) described in Eq.~\eqref{eq:adm} before being sent to the joint admittance block. The HapFIC algorithm (the joint-space HapFIC of Table~\ref{tab:modes}) within the joint admittance generates an offset $\delta$. This offset is added to the IK solution $q^{\mathrm{IK}}$—which is computed using $x^{\mathrm{adm}}$ as its input—to determine the admittance-modified joint angles $q^{\mathrm{adm}}$. These angles are subsequently sent as input to the Joint FICs (FIC$_{\mathrm{JS}}$) to compute the FIC control torques $\tau^J$. Finally, the gravity vector $g(q)$ is added to $\tau^J$ to generate the final actuator control torques $\tau_c$.}
  \label{fig:HPMP}
\end{figure}

\subsection{FIC Planner}\label{sec:planner}

The FIC planner is utilised to generate trajectories connecting via-points \cite{tiseo2021passive,tiseo2021fine}. It is implemented as an FIC governing a unit point mass moving within a viscous field. Consequently, the planner generates an acceleration profile that is numerically integrated to produce continuous velocity and position references \cite{tiseo2021passive,tiseo2021fine}. These resulting kinematic outputs are then strictly bounded to respect the specific physical limits of the robot hardware. Once a desired velocity is selected, the planner's parameters are automatically configured using principles from vibration theory. This ensures a critically damped convergence behaviour for both input via-points and continuous trajectories \cite{tiseo2021fine}.

\begin{equation}\label{eq:planner}
\left\{\begin{array}{ll}
     \ddot{x}=&\Gamma^I_c(\ddot{x}) - \mu \dot{x} \\\\
     \dot{x}=&\mathrm{max}\left(\dot{x}_\mathrm{min},\mathrm{min}\left(\dot{x}_\mathrm{max},\int_{t_0}^t\ddot{x}dt+\dot{x}(t_0)\right)\right)\\\\
     x=&\mathrm{max}\left(x_\mathrm{min},\mathrm{min}\left(x_\mathrm{max},\int_{t_0}^t\dot{x}dt+x(t_0)\right)\right)
\end{array}\right.
\end{equation}
where $t_0$ is the starting time, $t$  is the current time, and $\mu$ is the viscosity coefficient. $\dot{x}_\mathrm{max}$ and $x_\mathrm{max}$ are the upper for velocity and positions limits, respectively. $\dot{x}_\mathrm{min}$ and $x_\mathrm{min}$ are the lower velocity and position limits.  
\subsection{Admittance Control}\label{sec:admittance}
The admittance behaviour is divided into two distinct modes. Admittance accumulates the positional offset up to the hardware joint limits. In contrast, HapFIC bounds this offset using the $i^{th}\in[1,n]\subseteq\mathbb{N}$ and $j^{th}\in[n+1,k]\subseteq\mathbb{N}$ saturations (i.e., $x^i_{b}$ and $x^j_{b}$) of the FIC controller. Therefore, it implements a seemless trade-off between admittance and impedance behaviour \cite{hapfic2021,wen2023bimanual}.
In all other respects, Admittance and HapFIC (in both joint space and task space) operate identically according to the following admittance formulation. This implementation utilises a virtual viscosity field, determined by the coefficient $\mu_a$, alongside a desired virtual inertia $M$. By relying on this user-defined virtual inertia rather than the physical inertia of the robot, the admittance controller remains completely robust against kinematic singularities:
\begin{equation}\label{eq:adm}
\ddot{x}=M^{-1}\left(\mathrm{dz}_b\left(\tilde{\Gamma}\right) - \mu \dot{x}\right),\qquad \tilde{\Gamma} =\Gamma_d - \Gamma_c,
\end{equation}
where $\mathrm{dz}_b$ is a deadband with a band of $b$. The acceleration is then integrated in velocities and position as in the planner in equation \ref{eq:planner}.

\subsection{H-PMP: Nested Task-Space \& Joint-Space Admittance Controllers}\label{sec:hapfic}

The H-PMP architecture, illustrated in Fig.~\ref{fig:HPMP}, nests a task-space admittance or HapFIC controller (Admittance and HapFIC in task space, Table~\ref{tab:modes}) within a joint-space HapFIC (Table~\ref{tab:modes}). The task-space controller takes as input the desired pose ($x_{\mathrm{des}}$, represented as Cartesian coordinates plus a quaternion) and the external wrench ($w$). The wrench $w$ can either be provided directly by a sensor or estimated ($\hat{w}$) from the joint torques, which we source from the robots' APIs. The task-space controller generates the desired joint torques for the joint-space HapFIC as follows:
\begin{equation}
\label{eq:HPMP-TS}
\tau_{\mathrm{des}} = J^T w
\end{equation}
Meanwhile, the corresponding joint posture $q$ is generated via Inverse Kinematics (IK). The joint-space HapFIC then compares $\tau_{\mathrm{des}}$ with the net actuator torque (i.e., the measured actuator torque $\tau$ minus the gravity torque vector $g(q)$), which is compensated independently as depicted in Fig.~\ref{fig:HPMP}. Finally, the admittance output is passed to the Joint FICs to generate the torque command $\tau^J$, to which the gravity vector $g(q)$ is added to produce the final torque control signal $\tau_c$ for the actuators.

\section{Experimental Validation}\label{sec:eval}

The experiments validate the ability of the proposed H-PMP architecture to transition between impedance and admittance behaviours. All evaluations were conducted on a physical 7-DOF Kinova Gen3 robotic arm driven by the FIC library (Section~\ref{sec:system}) for Experiments~A, B, and C. The performance of the H-PMP is compared against three other FIC-based control architectures, each implemented with its own optimised tuning: (1) a Cartesian impedance controller with a planner~\cite{tiseo2022ik}, (2) a Task-Space HapFIC \cite{tiseo2021fine,wen2023bimanual} that sends commands directly to the actuators (i.e., $\tau_c=J^Tw_c +g(q)$), and (3) a Cartesian impedance controller with a planner sending commands to a joint-space HapFIC. This comparative analysis enables a rigorous assessment of the proposed controller's performance against current state-of-the-art FIC implementations in robotics. Finally, all experiments were implemented as extensions of FIC Studio, which is open-source and freely available on GitHub to facilitate reproducibility~\cite{ficinterface}.

The controllers are evaluated across three distinct tasks to assess their impedance and admittance behaviours. Experiment~A tests trajectory tracking performance. Experiment~B evaluates the admittance and impedance characteristics---as well as the transition between them---by having the robot make and lose contact with a wall. For this task, the admittance controllers are assigned a desired interaction force of zero, a condition that is abruptly altered when the obstacle suddenly collapses. Finally, Experiment~C compares the H-PMP and the Task-Space HapFIC during a learning-by-demonstration (LbD) task, utilising their admittance behaviour for kinesthetic teaching and their impedance behaviour for trajectory replay.

The hardware controller operates at a \SI{1}{\kilo\hertz} loop rate, interfacing with the robot via the vendor driver's torque interface. All evaluated controllers issue torque commands while simultaneously reading joint torques and positions. Additionally, the admittance controllers utilise the external wrench, which the library takes either from the robot's API or from an estimate based on the measured joint torques; on the Kinova Gen3, the estimate was used.

\begin{figure}[!t]
  \centering
  \includegraphics[width=\columnwidth]{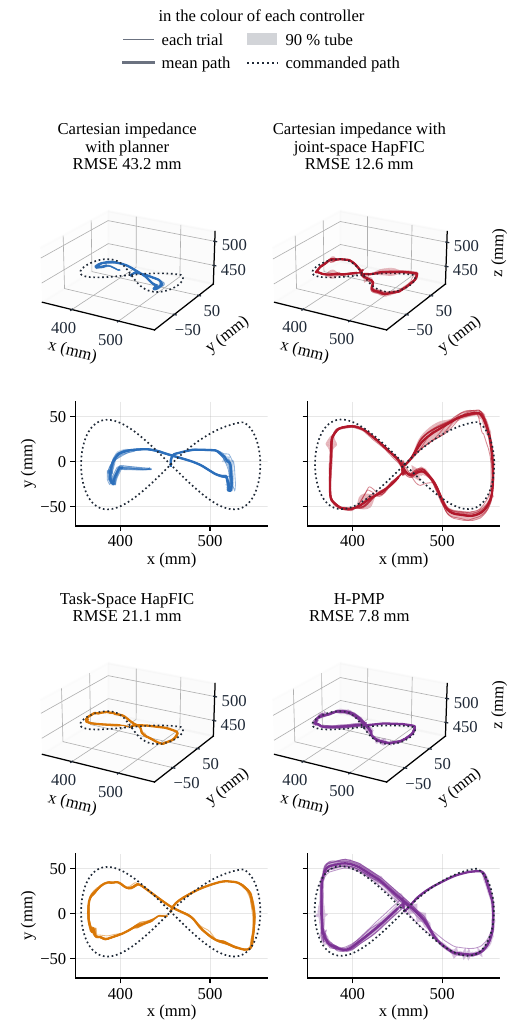}
  \caption{Trajectory tracking on the physical arm ($A=\SI{0.10}{\meter}$, \SI{20}{\second} period,
    one cycle per trial from the home pose, $n=10$ per controller): commanded path (dotted), every trial,
    the mean path and the \SI{90}{\percent} tube, in the colour of each controller. Each controller is shown
    in 3D with the horizontal plane below it: Cartesian impedance with planner and Cartesian impedance with
    joint-space HapFIC (top two rows), Task-Space HapFIC and H-PMP (bottom two rows).}
  \label{fig:track}
\end{figure}

\subsection{Trajectory Tracking}\label{sec:eval:track}
A lemniscate trajectory is utilised to evaluate how the controllers respond to continuous acceleration changes and direction inversions within the $xy$-plane, both of which pose significant challenges for interaction controllers. Let the robot's home position denote the centre of this trajectory. The commanded pose, which maintains a fixed end-effector orientation, is streamed at \SI{60}{\hertz}. The kinematic equations governing the lemniscate trajectory are defined as:
\begin{equation}\label{eq:lemniscate}
x=c_x+A\sin\omega t,\quad
y=c_y+A\sin\omega t\cos\omega t,\quad
z=c_z,
\end{equation}
where $A=\SI{0.10}{\meter}$ is the amplitude. The trajectory operates with a period of $T=2\pi/\omega=\SI{20}{\second}$. We evaluate the  tracking error as the instantaneous Euclidean distance between the measured and commanded end-effector positions, $d(t)=\lVert p(t)-p^{\mathrm{cmd}}(t)\rVert_2$. This distance metric is subsequently used to compute the mean absolute error (MAE), the root-mean-square error (RMSE), and the peak error. Furthermore, we evaluate the orientation error, calculated as the geodesic angle relative to the fixed target orientation, alongside the least-squares gain of the measured motion versus the commanded motion along the $x$-axis.

\begin{figure}[!t]
  \centering
  \includegraphics[width=\columnwidth]{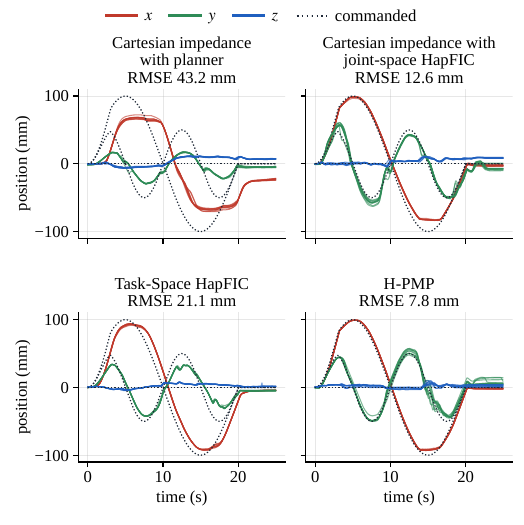}
  \caption{Trajectory tracking on the physical arm over time (the trials of Fig.~\ref{fig:track}): the
    $x$ (red), $y$ (green) and $z$ (blue) position of every trial from the centre $c$ of the lemniscate,
    commanded dotted, with the \SI{5}{\second} hold at the centre after the cycle.}
  \label{fig:track-t}
\end{figure}

The H-PMP tracked the lemniscate with an RMSE of \SI{7.80}{\milli\meter}, compared to \SI{21.11}{\milli\meter} for the Task-Space HapFIC (Table~\ref{tab:track}, Fig.~\ref{fig:track} and Fig.~\ref{fig:track-t}). The Cartesian impedance controller with joint-space HapFIC achieved an RMSE of \SI{12.56}{\milli\meter}, while the standard Cartesian impedance controller with a planner yielded \SI{43.19}{\milli\meter}. The H-PMP, Task-Space HapFIC, and Cartesian impedance with joint-space HapFIC followed the figure-eight trajectory with $x$-axis gains of 0.99, 0.97, and 0.94, respectively. In contrast, the Cartesian impedance with planner reached a gain of only 0.69, exhibiting a peak error of \SI{69.4}{\milli\meter} (compared to \SI{16.9}{\milli\meter}, \SI{32.2}{\milli\meter}, and \SI{20.9}{\milli\meter} for the first three controllers, respectively). The orientation error was \SI{3.33}{\degree} for the H-PMP and \SI{3.65}{\degree} for the Task-Space HapFIC, whereas the Cartesian impedance with joint-space HapFIC and the Cartesian impedance with planner maintained lower orientation errors of \SI{1.41}{\degree} and \SI{1.94}{\degree}, respectively. Across all configurations, the ten trials for each controller differed by at most \SI{2.04}{\milli\meter} in RMSE.

\begin{table}[!b]
\renewcommand{\arraystretch}{1.12}
\providecommand{\mci}[3]{\begin{tabular}[t]{@{}c@{}}#1\\[-1pt]{\scriptsize[#2, #3]}\end{tabular}}
\caption{Trajectory Tracking on the Arm: Mean [Range of the Ten Trials]}
\label{tab:track}
\centering
\footnotesize
\setlength{\tabcolsep}{2.8pt}
\begin{tabular}{@{}p{2.55cm}ccccc@{}}
\toprule
\textbf{Controller} & \textbf{MAE} & \textbf{RMSE} & \textbf{Peak} & \textbf{Ori.} & \textbf{$x$ gain} \\
& (mm) & (mm) & (mm) & ($^{\circ}$) & \\
\midrule
Cartesian impedance with planner & 41.4 & \mci{43.19}{42.65}{43.72} & \mci{69.4}{69.0}{69.8} & 1.94 & 0.69 \\
Cartesian impedance with joint-space HapFIC & 11.8 & \mci{12.56}{11.81}{13.65} & \mci{20.9}{20.3}{22.2} & 1.41 & 0.94 \\
Task-Space HapFIC & 20.3 & \mci{21.11}{21.00}{21.22} & \mci{32.2}{31.8}{32.4} & 3.65 & 0.97 \\
H-PMP & 7.3 & \mci{7.80}{7.06}{9.10} & \mci{16.9}{15.5}{18.8} & 3.33 & 0.99 \\
\bottomrule
\end{tabular}
\end{table}

Fig.~\ref{fig:track} and Fig.~\ref{fig:track-t} show that the H-PMP achieves the best tracking performance in terms of error; however, its trajectory is less smooth compared to the purely Cartesian controllers (Cartesian Impedance and HapFIC). It does still yield a smoother trajectory than a Cartesian Impedance controller combined with a joint-space HapFIC. The oscillatory motion introduced in the H-PMP trajectory can therefore be attributed to the joint-space HapFIC trading tracking accuracy for torque tracking. This issue can be mitigated by utilising the robot's dynamic model to enhance the desired torque estimation or by introducing online tuning to adapt the controller's behaviour.

\subsection{Wall Press and Release}\label{sec:eval:wall}
The experimental setup consists of a vertical plate on a quick-release mount positioned beside the arm, facing the $x$-axis of the base frame (Fig.~\ref{fig:summary}). The robot controls the physical interaction with the rigid wall until the experimenter activates the quick release, causing the wall to collapse and generating a sudden loss of end-effector contact. Such a task is notoriously difficult for traditional interaction controllers. An admittance approach can effectively regulate the interaction force, but the sudden loss of contact can drive the controller into instability. Conversely, an impedance controller handles the loss of contact safely but only allows for indirect control of the interaction force via feed-forward estimation. Therefore, this serves as an excellent task to validate the claim that the proposed architecture successfully seamlessly combines admittance and impedance behaviours.

\begin{table}[!t]
\renewcommand{\arraystretch}{1.05}
\caption{Wall Press and Release on the Arm, Expected Maximum Force \SI{20}{\newton}: Mean [Range]}
\label{tab:wall}
\centering
\scriptsize
\setlength{\tabcolsep}{2.4pt}
\begin{tabular}{@{}p{3.8cm}rrrr@{}}
\toprule
\textbf{Controller} & $n$ & \textbf{$F_{\mathrm{hold}}$} & \textbf{shift} & \textbf{past} \\
 & & (N) & (mm) & (mm) \\
\midrule
\parbox[t]{3.8cm}{\raggedright Cart.\ imp.\ with joint-space HapFIC} & 2 & 16.9  & 0.0 & 68.7 \\
\parbox[t]{3.8cm}{\raggedright Task-Space HapFIC} & 1 & 4.6 & $-$50.0 & 36.3 \\
H-PMP & 2 & 12.6 & $-$50.0 & 23.4 \\
\bottomrule
\end{tabular}

\smallskip
\parbox{\columnwidth}{\scriptsize Expected force: the task-space law at \SI{100}{\milli\meter} past the wall face.
  $F_{\mathrm{hold}}$: contact force at the end of the hold; shift: set-point shift at the end of the hold; past: largest
  distance past the wall face after the release.}
\end{table}

The robot makes contact with the wall by attempting to reach a desired state located \SI{60}{\milli\meter} beyond the wall along the $x$-axis. Following a \SI{1}{\second} baseline phase, the target moves further along the $x$-axis towards the wall at \SI{10}{\milli\meter\per\second} until it reaches \SI{100}{\milli\meter} beyond the face, at which point the task-space law commands an expected contact force of \SI{20}{\newton}. The target is then held. Between \SI{4}{\second} and \SI{14}{\second} into this hold phase, the operator pulls the release. The plate drops away while the commanded target remains fixed; the arm's response is then recorded for an additional \SI{5}{\second}. Because no external force sensor is fitted, the contact force relies on the controllers' estimation of the external wrench derived from the joint torques. This is measured along the press axis and tared prior to each press. Three architectures are compared over two presses each (one for the Task-Space HapFIC): Task-Space HapFIC, Cartesian impedance with joint-space HapFIC, and H-PMP. 

\begin{figure}[!t]
  \centering
  \includegraphics[width=\columnwidth]{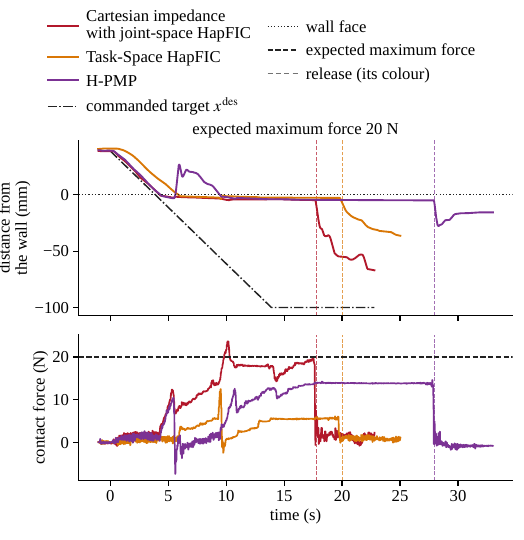}
  \caption{Wall press and release on the physical arm at an expected maximum force of \SI{20}{\newton} (first press
    of each controller; H-PMP and Cartesian impedance with joint-space HapFIC with the contact profile). Top: the
    distance from the wall (zero at the touch, negative past the face; the wall face dotted) and the commanded target
    $x^{\mathrm{des}}$ (dash-dot), which ramps \SI{100}{\milli\meter} past the face and is then held; bottom: the contact force
    estimated from the joint torques, with the expected maximum force (dashed). The vertical dashed lines mark each
    controller's release, in its colour.}
  \label{fig:wall}
\end{figure}

The Cartesian impedance controller with joint-space HapFIC sustained the highest contact force and interaction torque (Table~\ref{tab:wall}, Fig.~\ref{fig:wall}), registering \SI{16.86}{\newton} just prior to the release. Because its set-point did not shift, it travelled \SI{68.7}{\milli\meter} past the original wall face immediately following the release. In contrast, the H-PMP shifted its set-point by \SI{50.0}{\milli\meter} and maintained a contact force of \SI{12.59}{\newton}, safely travelling only \SI{23.4}{\milli\meter} past the face upon release. The Task-Space HapFIC also shifted its set-point by \SI{50.0}{\milli\meter} but held a lower force of \SI{4.59}{\newton}, travelling \SI{36.3}{\milli\meter} past the face. Notably, during its first press, the H-PMP rebounded from the wall once before settling into stable contact.

\begin{figure}[!t]
  \centering
  \includegraphics[width=\columnwidth]{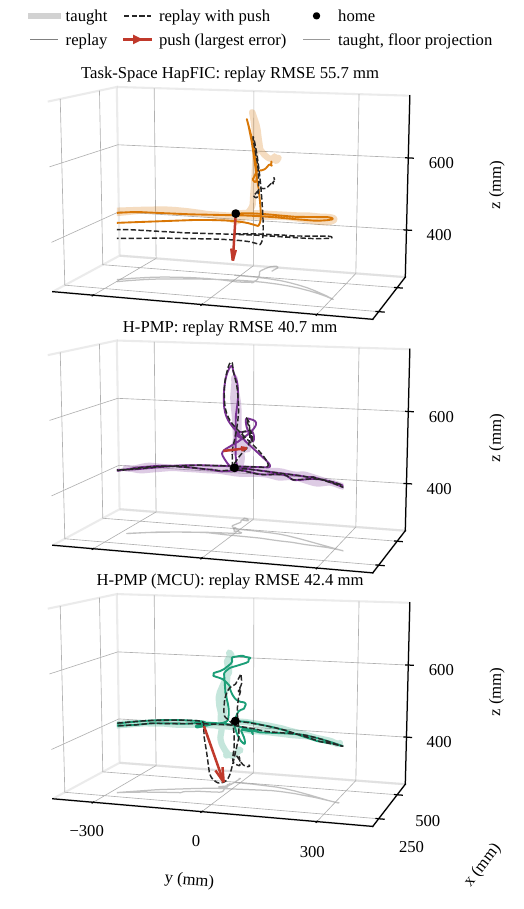}
  \caption{Learning by demonstration on the physical arm (one teaching per setup, by hand; bottom: the
    H-PMP with its joint-space HapFIC on the MCU): the taught path (wide, light), the replay (solid) and
    the replay with the hand push (dashed) in 3D, over the taught movement.}
  \label{fig:lbd}
\end{figure}

\begin{figure}[!t]
  \centering
  \includegraphics[width=\columnwidth]{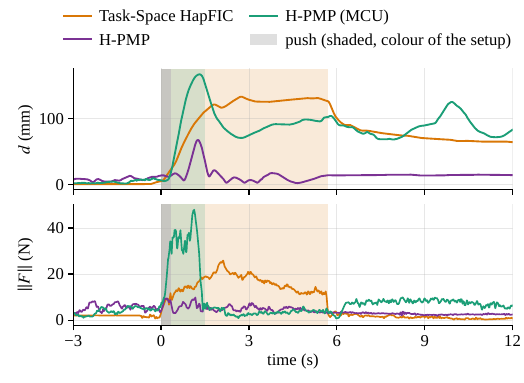}
  \caption{Learning by demonstration on the physical arm, about the hand push of Fig.~\ref{fig:lbd}. Top: $d$,
    the end-effector's distance from the unpushed replay at the same replay time; bottom: $\lVert F\rVert$, the force
    on the end-effector during the pushed replay, estimated from the joint torques (\SI{50}{\milli\second} mean); both against time from
    the start of the push ($t=0$). The push, located in the recorded force estimate, is shaded in the colour of
    its setup.}
  \label{fig:lbd-t}
\end{figure}

The wall press and release data in Fig. \ref{fig:wall} indicates that the H-PMP's force-tracking performance falls between that of the purely Cartesian HapFIC and the Cartesian controller with a joint-space HapFIC, while outperforming both in recovering from contact loss. The end-effector detachment observed upon making contact with the wall suggests an initial overcompensation. This is subsequently counteracted when the HapFIC reaches the maximum allowed displacement of its desired state under the admittance behaviour. When contact is lost, the H-PMP exhibits the smoothest transition from admittance to impedance behaviour. These results further confirm that performance could be enhanced by introducing online parameter tuning, enabling the controller to adapt to the task at hand.

\subsection{Learning by Demonstration}\label{sec:eval:lbd}

The learning-by-demonstration (LbD) task evaluates the ability to seamlessly switch from admittance to impedance behaviour during physical interaction with a user. The user relies on the admittance behaviour to kinesthetically demonstrate a trajectory to the robot, which must then be replayed by the controller exploiting its impedance behaviour. The only baseline configuration that can be fairly compared with the H-PMP for this task is a combination of a Cartesian admittance controller (for the teaching phase) and the Task-Space HapFIC (for the replay phase). All controller tunings remain identical across both phases.

After homing, the teaching controller operates in either the baseline Cartesian admittance mode or the H-PMP task-space admittance mode. The operator moves the end-effector from the home pose to the left, to the right, up, down and out, as shown in Fig.~\ref{fig:lbd}. The arm is then homed again, and the taught trajectory is replayed using its original timing in either the Task-Space HapFIC baseline or the H-PMP operating in HapFIC mode in the Task-Space. A second replay is subsequently performed with an introduced physical perturbation to evaluate the perturbation rejection capabilities of the controllers' impedance behaviours. The H-PMP was also run with its joint-space HapFIC on a microcontroller (MCU): the STM32H5 of an Arduino Ventuno Q runs it at \SI{1}{\kilo\hertz} and drives the arm through an Ethernet shield, while the task-space controller runs on the board's Linux processor and reaches the MCU over the board's internal link at about \SI{95}{\hertz}. The task-space controller could not be integrated into the MCU due to its dependency on Pinocchio. However, because the current control loop and robot communication require only about \SI{100}{\micro\second}, there is ample computational capacity to include the task-space controller alongside analytical or geometrical kinematic models directly on the MCU.

\begin{table}[!t]
\renewcommand{\arraystretch}{1.08}
\caption{Learning by Demonstration on the Arm (One Teaching per Setup)}
\label{tab:lbd-hw}
\centering
\scriptsize
\setlength{\tabcolsep}{2.5pt}
\begin{tabular}{@{}lccc@{}}
\toprule
 & \textbf{\begin{tabular}[b]{@{}c@{}}Task-Space\\HapFIC\end{tabular}} & \textbf{H-PMP} & \textbf{\begin{tabular}[b]{@{}c@{}}H-PMP\\(MCU)\end{tabular}} \\
\midrule
Teaching speed, mean (\si{\milli\meter\per\second}) & 150 & 144 & 75 \\
Replay RMSE over the movement (mm) & 55.7 & 40.7 & 42.4 \\
Replay error, peak (mm) & 119.0 & 92.8 & 87.6 \\
Peak perturbation force $F_{\mathrm{p}}$ (N) & 27.8 & 11.3 & 51.3 \\
Peak deviation $d_{\max}$ (mm) & 132 & 67 & 167 \\
Residual deviation $d_{\mathrm{r}}$ (mm) & 77 & 4.6 & 85 \\
Compliance $C$ (\si{\milli\meter\per\newton}) & 4.8 & 6.0 & 3.3 \\
\bottomrule
\end{tabular}

\smallskip
\parbox{\columnwidth}{\scriptsize $d$: Euclidean distance of the end-effector from the unperturbed replay at
  equal replay time; $\lVert F\rVert$: end-effector force estimated from the joint torques. $F_{\mathrm{p}}$: peak
  of $\lVert F\rVert$ during the perturbation; $d_{\max}$: peak of $d$ over the perturbation and the following
  \SI{2}{\second}; $d_{\mathrm{r}}$: $d$ at \SI{2}{\second} after the perturbation; $C = d_{\max}/F_{\mathrm{p}}$.}
\end{table}

Over the taught movement, the H-PMP replayed its taught path with an RMSE of \SI{40.7}{\milli\meter} and a peak error of \SI{92.8}{\milli\meter}, compared to \SI{55.7}{\milli\meter} and \SI{119.0}{\milli\meter} for the Task-Space HapFIC (Fig.~\ref{fig:lbd}, Table~\ref{tab:lbd-hw}), with both paths taught at a mean hand speed of \SI{144}{\milli\meter\per\second} and \SI{150}{\milli\meter\per\second}, respectively. During the perturbation tests, both controllers safely yielded to the pushes. We define the compliance as the peak error, the end-effector's distance from the unperturbed replay, over the push and the \SI{2}{\second} after it, divided by the peak force of the push, both located in the recorded data. Under a hand push of \SI{11.3}{\newton}, the H-PMP's end-effector moved up to \SI{67}{\milli\meter} from the unperturbed replay trajectory (Fig.~\ref{fig:lbd-t}), yielding a compliance of \SI{6.0}{\milli\meter\per\newton}. Similarly, under a push of \SI{27.8}{\newton}, the Task-Space HapFIC deviated by up to \SI{132}{\milli\meter}, a compliance of \SI{4.8}{\milli\meter\per\newton}. With its joint-space HapFIC on the MCU, the H-PMP replayed a path taught at \SI{75}{\milli\meter\per\second} with an RMSE of \SI{42.4}{\milli\meter} and a peak error of \SI{87.6}{\milli\meter}; under a push of \SI{51.3}{\newton}, its end-effector moved up to \SI{167}{\milli\meter} (\SI{3.3}{\milli\meter\per\newton}) and was \SI{85}{\milli\meter} from the unperturbed replay \SI{2}{\second} after the push.

The LbD experimental data in Fig. \ref{fig:lbd} confirm the superiority of the H-PMP over the Cartesian controllers in transitioning from admittance to impedance behaviour, generating a more accurate replica of the trained trajectory.

\section{Conclusion}\label{sec:conclusion}

We presented an evaluation of the H-PMP against other FIC controllers, demonstrating that it offers a balanced compromise between impedance and admittance behaviours. This enables the robot to successfully track trajectories while controlling interaction forces with the environment. The H-PMP architecture is computationally efficient and could be executed in its entirety on the MCU, which has a maximum power consumption of approximately \SI{400}{\milli\watt}~\cite{stm32h5f}, compared to the \SIrange{15}{17}{\watt} drawn by the Arduino Ventuno Q board, built around the highly efficient Qualcomm Dragonwing IQ-8275~\cite{ventunoqpower}. Furthermore, the data suggest that performance could be enhanced by exploiting the FIC's real-time capabilities to introduce online tuning, thereby implementing an adaptive architecture. Finally, we also introduced a FIC Suite that enables the deployment of all available FIC architectures in both MuJoCo simulations and on physical robots. Future work will focus on developing adaptive tuning methods to enable the online optimisation of FIC performance.

\bibliographystyle{IEEEtran}
\bibliography{refs}

\end{document}